\documentclass[conference]{IEEEtran}
\IEEEoverridecommandlockouts
\usepackage[T1]{fontenc}
\usepackage{times}
\usepackage{microtype} % Helps with font embedding and spacing
\usepackage{cite}
\usepackage{amsmath,amssymb,amsfonts}
\usepackage{algorithm}
\usepackage{algpseudocode}
\usepackage{graphicx}
\usepackage{textcomp}
\usepackage{multirow}
\usepackage{csquotes}
\usepackage{listings}
\usepackage{lettrine}
\usepackage{array}
\usepackage{stfloats}
\usepackage{url}
\usepackage{verbatim}
\usepackage{upgreek}
\usepackage{siunitx}
\usepackage{adjustbox}
\usepackage[inline]{enumitem}
\usepackage{xcolor}
\usepackage{threeparttable}
\usepackage[caption=false]{subfig}
\usepackage{booktabs}

\usepackage{tabularx}

\begin{document}

\title{Battery State of Health Estimation Using LLM Framework}

\makeatletter
\newcommand{\linebreakand}{%
  \end{@IEEEauthorhalign}
  \hfill\mbox{}\par
  \mbox{}\hfill\begin{@IEEEauthorhalign}
}
\makeatother

\author{%

    \IEEEauthorblockN{Aybars Yunusoglu\IEEEauthorrefmark{1}}
    \IEEEauthorblockA{\textit{Purdue University} \\
    West Lafayette, USA \\
    ayunusog@purdue.edu \\
    \IEEEauthorrefmark{1}Corresponding author}
    \and
    \IEEEauthorblockN{Dexter Le\IEEEauthorrefmark{1}}
    \IEEEauthorblockA{\textit{Drexel University} \\
    Philadelphia, USA \\
    dql27@drexel.edu \\
    \IEEEauthorrefmark{1}Corresponding author}
    \and
    \IEEEauthorblockN{Murat Isik}
    \IEEEauthorblockA{\textit{Stanford University} \\
    Stanford, USA \\
    misik@stanford.edu}
    
    \linebreakand % Force authors to next row

    \IEEEauthorblockN{Karn Tiwari}
    \IEEEauthorblockA{\textit{Indian Institute of Science, Bangalore} \\
    Bengaluru, India \\
    karntiwari@iisc.ac.in}
    \and
    \IEEEauthorblockN{I. Can Dikmen}
    \IEEEauthorblockA{\textit{Temsa Research \& Development Center} \\
    Adana, Turkey \\
    can.dikmen@temsa.com}

 }   
\maketitle

\begin{abstract}

Battery health monitoring is critical for the efficient and reliable operation of electric vehicles (EVs). This study introduces a transformer-based framework for estimating the State of Health (SoH) and predicting the Remaining Useful Life (RUL) of lithium titanate (LTO) battery cells by utilizing both cycle-based and instantaneous discharge data. Testing on eight LTO cells under various cycling conditions over 500 cycles, we demonstrate the impact of charge durations on energy storage trends and apply Differential Voltage Analysis (DVA) to monitor capacity changes (dQ/dV) across voltage ranges. Our LLM model achieves superior performance, with a Mean Absolute Error (MAE) as low as 0.87\% and varied latency metrics that support efficient processing, demonstrating its strong potential for real-time integration into EVs. The framework effectively identifies early signs of degradation through anomaly detection in high-resolution data, facilitating predictive maintenance to prevent sudden battery failures and enhance energy efficiency.

\end{abstract}


\section{Introduction}
Lithium titanate battery cells are used across various industries and preferred for electric vehicles due to LTO performances. The LTO performances range from faster charging times and lifetime when compared to other lithium-ion battery solutions \cite{nemeth2020, le2024multimodal}. Despite LTOs being expensive and having low energy density, the potential for these batteries is investigated as \cite{dang2023} have concluded 54.9\% fuel savings in hybrid-electric heavy-duty vehicles when compared to diesel trucks. This underlines the efficiency, potential, and durability of LTO battery cells, making it a beneficial battery cell for this context.

SoH estimation is crucial in the context of electric vehicles, as the estimation is responsible for ensuring safety through efficient use of the battery \cite{nakano2024, unger2016battery} and AI hardware performance \cite{isik2024accelerating, isik2023survey}. To provide a safe and reliable SoH estimation, we are using multiple methodologies that optimize estimations' accuracy. 

The SoH estimation methods have been developed over the years with the rapid development of Artificial Intelligence and Machine Learning. In 2015 \cite{lin2015review} reviewed SoH estimation methods in lithium ion batteries for EVs. They used statistical analysis (analytical models) and microscopy measurements to observe electrodes and electrolytes. In 2020, \cite{noura} and \cite{aloisio2020machine} noted that internal resistance, impedance measurements, and IC/DC curves are frequently used in estimating SoH of batteries in the context of hybrid electric vehicles. In 2023, \cite{lixu2023} and \cite{patil2023lithium} implemented LSTM (long short-term memory network) to predict the future behavior of battery health. This depicts the involvement of AI/ML through years for use in estimating SoH/SoC \cite{zheng4910174adaptive, cheng2024shield, bian2024hybrid, yildiran2024state}. This paper introduces an LLM framework to estimate the SoH and RUL of LTO battery cells through anomaly detection. Additionally, the LLM framework utilizes DVA to assist in the estimation of SoH for LTOs.

The primary contributions of this paper center around the development and evaluation of a unified multimodal LLM framework aimed at advancing intelligent transport systems. The main contributions of this paper highlight a novel approach for estimating battery State of Health (SoH) and Remaining Useful Life (RUL) using a multimodal LLM framework. Our contributions include:

\begin{itemize} 
\item Development of a multimodal LLM framework optimized for precise SoH and RUL predictions in battery systems, leveraging transformer-based modeling to handle cycle-based and instantaneous data. \item Introduction of Differential Voltage Analysis (DVA) within the LLM framework, enhancing feature extraction for more accurate health assessments of lithium titanate (LTO) cells. \item Extensive evaluation against traditional and advanced predictive models, demonstrating the framework’s improved accuracy with MAE values as low as 0.87\%, and analyzing its potential for integration in EV predictive maintenance. 
\end{itemize}

The paper is organized as follows: \textbf{Section 2} presents the background and existing methodologies for battery health estimation, including recent developments in transformer-based SoH and RUL prediction. \textbf{Section 3} details the dataset, battery testing conditions, and data preparation process, along with a description of the LLM framework used in this study. \textbf{Section 4} outlines the specific machine learning algorithms and evaluation metrics applied to assess model performance. \textbf{Section 5} provides an analysis of the proposed LLM framework's performance and accuracy in battery SoH and RUL estimation compared to traditional methods. \textbf{Section 6} concludes with potential directions for future work, including improvements to processing speed and model scalability.

\section{Background}
\subsection{Transformer-based SoH and SoC estimation}
SoH/SoC battery estimation is vital for controlling electric vehicles and securing power responses \cite{sundén2019hydrogen}. Consequently, many researchers have improvised accuracy for estimation methods. For instance \cite{feng2024gpt4battery}, developed a relatively accurate LLM-driven framework for SoH estimation of LIBs called GPT4 Battery. It has been reported that the MAE average of four well-known datasets (CALCE, SANYO, KOKAM and PANASONIC) was 2.17\% under zero-shot settings. As a result, this emphasizes the improvement and accuracy of SoH estimation using LLM-driven frameworks. This inspired us to create a framework that detects battery degradation. 

A survey study \cite{guirguis2024} listed various advancements in Transformer-based architectures for estimating SoH/SoC of LIBs by reviewing previous research and 15 real-world related well-known datasets. A table summary of SoC estimation studies in the paper records that \cite{hannan2021} 0.44\% MAE and 0.9\% RMSE are achieved by using a Transformer with SSL methodology and LG18650HG2 dataset under constant temperature conditions. The most accurate state of health (SoH) estimation for lithium-ion batteries is achieved using the exponential smoothing transformer (SGEformer), which attained a MAE of 0.01\% with datasets from NASA and CALCE. 

\subsection{DVA integration for SoH Estimation}
DVA implementation is frequently used in SoH estimation to analyze structural transformation and assist feature extraction, and it has received significant attention in recent years for estimation purposes \cite{demirci2024review, oji2021data}. \cite{xiafei2023} used the DVA method and incremental capacity analysis (ICA) to predict the RUL and SoH of lithium-ion batteries. In this context, DVA is used to construct a clean DV curve that can help indicate the health statement of batteries since the peaks of the DV curve create regions that show the capacity of phase transitions, ultimately assisting in estimating battery age. The result of the experiment shows that the RMSE of the CALCE battery is 0.0006, indicating a strong reliability of using a DVA model. 
However, it is important to consider that this experiment does not manage noises entirely, hence uses voltage reconstruction which may cause accuracy and resolution limitations.

\section{Mathematical Expression}

\subsection{Differential Voltage Analysis (DVA)}
Differential Voltage Analysis (DVA) is integrated into the LLM framework to enhance feature extraction by detecting subtle variations in battery performance. DVA evaluates the differential capacity (\(dQ\)) per unit voltage change (\(dV\)), providing insights into the battery’s internal state and capacity degradation over cycles. Mathematically, DVA can be expressed as:

\begin{equation}
\frac{dQ}{dV} = \lim_{\Delta V \to 0} \frac{Q(V + \Delta V) - Q(V)}{\Delta V}
\end{equation}

The charge capacity is calculated as a function of voltage and this illustrates the phase transition as well as the structural changes within the battery. When the voltage is between 2.25V and 2.30V, there is a decrease in energy storage from 40 mAh in the 50th cycle to 28 mAh in the 500th cycle. Such a decrease shows that the battery is aging. The LLM framework enables the detection of degradation patterns to predict SoH and RUL with higher accuracy.

\subsection{SoH and RUL Estimation}
LLM uses transformer layers and temporal attention mechanisms to predict the batteries' State of Health (SoH). Thus, the long-term dependencies in battery data are identified. This ensures more accurate capacity forecasting. The following formula can be used to calculate SoH:

\begin{equation}
\text{SoH}(\%) = \left( \frac{\text{Current Charge Capacity (mAh)}}{\text{Nominal Capacity (mAh)}} \right) \times 100
\end{equation}

Current Charge Capacity is the measured capacity of the battery at a given cycle. Nominal Capacity is the rated capacity of the battery in the new condition. The battery health is represented in terms of percentage in this formula. 
The Remaining Useful Life (RUL) is estimated with non-linear degradation patterns by the LLM framework via polynomial regression. The cycle is estimated to predict the end of the useful life cycle for the battery applications and this is done when the charge capacity falls under 80\% of the nominal capacity. The solution is provided for cycle (C\_{\text{end}}) to calculate the threshold:

\begin{equation}
\text{SoH}(C_{\text{end}}) = 80
\end{equation}

The LLM models this using polynomial regression on the SoH curve:

\begin{equation}
\text{SoH}(C) = aC^2 + bC + c
\end{equation}

Here, a, b and c are coefficients obtained by fitting the model to past SoH data. C, on the other hand, represents the cycle number. This model can predict the approximate cycle for the end of the battery’s useful life when solved for (C\_{\text{end}}). It is possible to create a predictive maintenance tool using DVA-based extraction and LLM-driven SoH and RUL estimation. Such prediction can improve the reliability and efficiency of battery-powered transportation systems.

\section{Methods}

This paper develops a transform-based LLM framework for estimating the SoH and Remaining Useful Life (RUL) of lithium titanate (LTO) batteries. The predictive performance can be improved with LLM’s capability to extract patterns from complex datasets. This can be done by combining the cycle-based capacity data and instant discharge measurement.

\subsection{Dataset}
The dataset used in this paper includes the battery ageing data from eight LTO battery cells under different cycle conditions. For each cycle, the total charge capacity (Cap\_Chg, measured in mAh) and discharge capacity were recorded. The SoH is calculated as the ratio of the current capacity to the nominal capacity. As soon as the SoH dropped below 80\%, the battery was considered to have reached the end of its useful life, in line with industry standards for electric vehicles. During the discharge process, instantaneous voltage measurements were taken at specific cycle intervals in addition to cycle-based data, providing high-resolution insights into real-time battery behavior and detecting early degradation signals that are released in \cite{4bqx-gt92-24}.

\begin{figure*}
    \graphicspath{ {D:\Stack} }
    \center \includegraphics[width=1\textwidth]{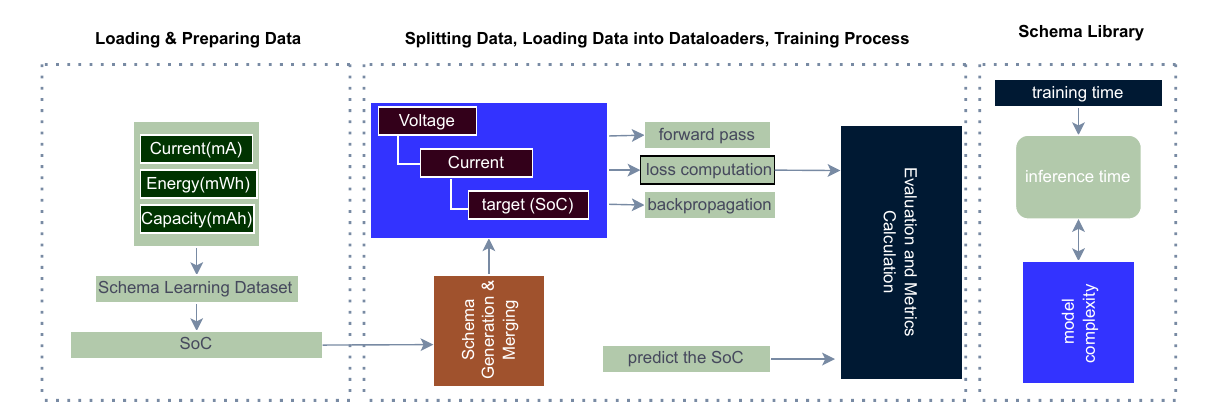}
    \caption{Pipeline for State of Charge (SoC) Prediction Using Multi-Input Regression with Schema-Based Data Processing and Model Evaluation}
    \label{test1}
\end{figure*}

An integrated pipeline for predicting the State of Charge (SoC) of batteries is presented in Figure \ref{test1} using a multi-input regression framework that integrates data processing, schema learning, model training, and evaluation. Pipelines are organized into stages, each contributing to the overall process. Schema Learning Datasets involve gathering and structuring raw battery data from a variety of sources. The essential relationships between Voltage (V), Current (mA), Capacity (mAh), and Energy (mWh) are captured in a schema learning dataset that supports feature extraction for SoC prediction. The predictive model is based on this dataset.

The Data Handling and Preparation phase involves data loading and preprocessing, which involves tasks such as cleaning, normalization, and feature extraction. The dataset is processed and divided into two subsets namely the training subset and the testing subset. Both of these subsets were loaded to PyTorch DataLoaders to ensure efficient batching, shuffling and data optimization when handling the model training. The Model Training Process consisted of three stages. The first stage is forward pass and at this stage, input data propagates through the network layers to produce initial SoC predictions. The second stage is loss computation and backpropagation and at this stage, the discrepancy between predicted and actual SoC values is computed using a loss function and the parameter updates are guided to minimize the error across successive epochs. The third stage is performance tracking and at this stage, metrics including the training time, model complexity and loss reduction are recorded to evaluate the model's efficiency.
This component incorporates patterns observed during the training to iteratively refine the SoC prediction schema and this improves the model’s generalizability. The model is evaluated in the Prediction and Evaluation stage by predicting the SoC values on the test dataset after training. The prediction accuracy is quantified by calculating performance metrics including the mean squared error (MSE), R2 score and MAE. At the same time, computational metrics including the inference time and model complexity which is determined by parameter count gave insight into the model’s efficiency and feasibility for real-time applications. The Schema Library and Final SoC Prediction Reporting coalesced learned structures and document the relationships between input features such as voltage, current and capacity as well as SoC predictions. The compiled results with final predictions and evaluation metrics can be an important tool for an overview of the model’s accuracy, computational performance and applicability.

\section{Model}

The proposed framework based on a transformer-based LLM can process the structured cycle data and unstructured instant discharge data. The LLM was fine-tuned to detect the complex dependencies and trends as well as the time-series data from charge-discharge cycles and voltage measurements. LLM adopts normalized charge and discharge capacity data over multiple cycles and metadata, including the temperature and cycle rate, to conduct a high-resolution trend analysis. The long-term dependencies across cycles can be captured by the architecture since it is possible to learn from small variations in the data. LLM predicts the remaining usage life of the battery and computes an accurate SoH prediction by utilizing the temporal patterns and cycle trends. The LLM framework utilizes instant discharge data to identify anomalies. The model marks any anomalies that might be a potential fault or unexpected degradation. These anomalies are marked based on the deviations from the expected voltage patterns. Outliers are identified, and a feedback mechanism is applied to refine the predictions.

\begin{figure*}
    \graphicspath{ {D:\Stack} }
    \center \includegraphics[width=1\textwidth]{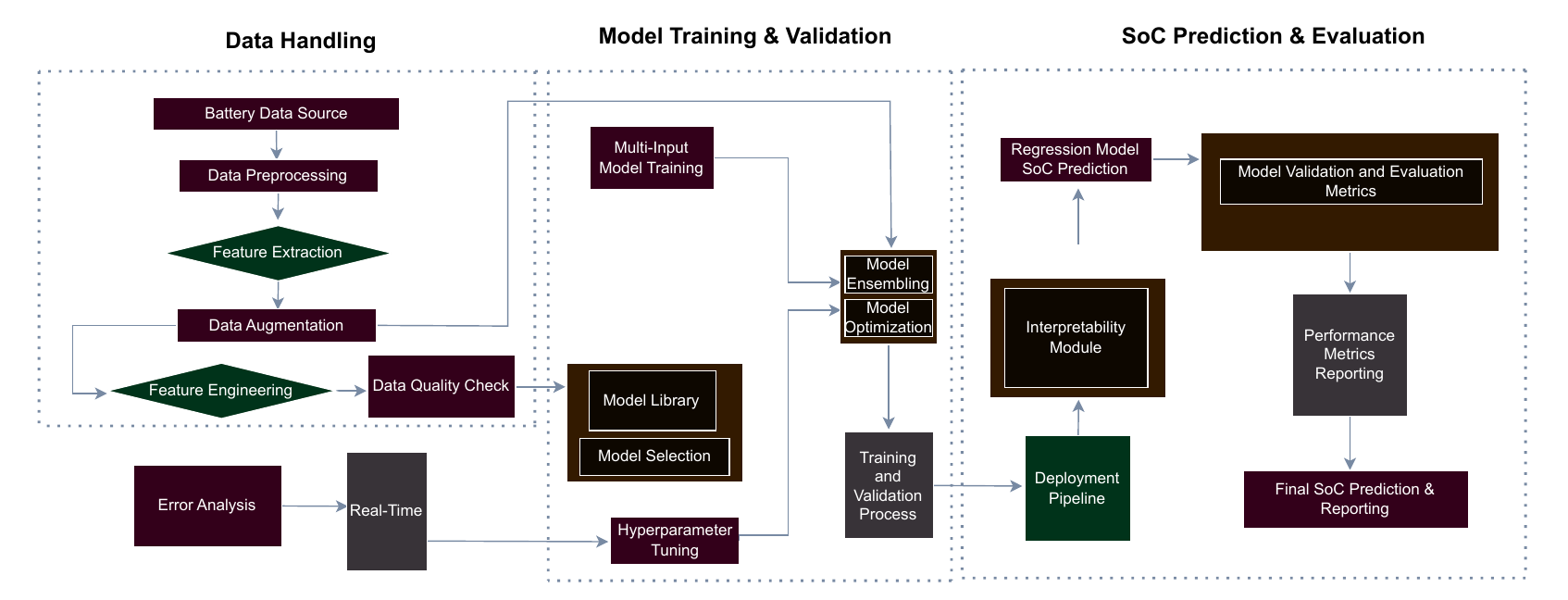}
    \caption{End-to-End Pipeline for State of Charge (SoC) Prediction: Data Handling, Model Training \& Validation, and Evaluation }
    \label{test2}
\end{figure*}

Figure \ref{test2}, an end-to-end pipeline for the State of Charge (SoC) prediction of a battery is given. The main focus here is the sequential processes involved in the data handling, model training and validation as well as SoC prediction for performance evaluation. Data Handling is the first stage of the pipeline. It gathers data from different Battery Data Sources and normalizes this data. This pre-processing stage is necessary for high-quality input data for model training. The relevant parameters are identified by the Feature Extraction. In addition to that, the Feature Engineering and Data Augmentation refine the data to increase the model's robustness and generalizability. Moreover, the dataset integrity and reliability are provided by the Data Quality Check to decrease the risk of errors during the downstream analysis.
After the structured training process, Model Training and Validation with machine learning techniques are implemented. Multi-Input Model Training utilizes multiple input features for SoC prediction and a comprehensive understanding of how the battery behaves. Hyperparameter tuning is applied to optimize the model parameters for enhanced performance. Training and Validation are applied to refine model accuracy via iterative testing on validation sets. Moreover, an Error Analysis phase is applied to set up the model for deployment in a dynamic environment with the purpose of identifying and correcting the inaccuracies. Data Handling is the first stage of a pipeline. At this stage, data is collected, cleaned and normalized from different Battery Data Sources. This pre-processing is necessary to ensure high-quality input data for model training. For identifying the relevant parameters, Feature Extraction is applied. Feature Engineering and Data Augmentation are applied to enhance the model’s robustness and generalizability. Moreover, Data Quality Check is applied for the dataset integrity and reliability to minimize the risk of errors across downstream analysis. At the Model Training and Validation stage, machine learning techniques were applied according to a structured training process. Multiple input features are utilized by the Multi-Input Model training to predict SoC behavior. This enables us to understand the battery behavior thoroughly. The hyperparameters are tuned to enhance the performance and the model’s accuracy is refined via iterative testing with training and validation. Additionally, the Error Analysis phase of the model identifies and corrects the inaccuracies before deploying the model in a dynamic environment. The framework was implemented by PyTorch and the framework was fine-tuned by considering the past battery datasets to achieve generalization across the different cycling rates and operating conditions. The predictive accuracy to prevent overfitting was assessed with performance metrics such as MAE and Root Mean Square Error (RMSE).

The data loading and preparation script begins by scanning a specified directory for Excel files related to battery data. The function load\_and\_prepare\_data reads each Excel file, dynamically identifying key columns such as Cap\_Chg(mAh), Cap\_DChg(mAh), and Voltage(V) using a helper function. Each row in the sheet is summarized as a text input by concatenating values from relevant columns, with the charge or discharge capacity column assigned as the regression label. After processing all valid files, the data is concatenated into a single pandas DataFrame containing text-label pairs. The text-label pairs are split into training and testing datasets using an 80-20 split, with tokenization handled by the BERT tokenizer from the transformers library. This tokenizer converts text data into tokenized input suitable for the model.

For structured data handling, the BatteryDataset class inherits from torch.utils.data.Dataset to structure the data for PyTorch’s DataLoader. Each text input is encoded using the BERT tokenizer, with padding or truncating to a maximum length of 128 tokens. The DataLoader generates data batches for efficient training and evaluation.

\begin{figure*}
    \graphicspath{ {D:\Stack} }
    \center \includegraphics[width=0.76\textwidth]{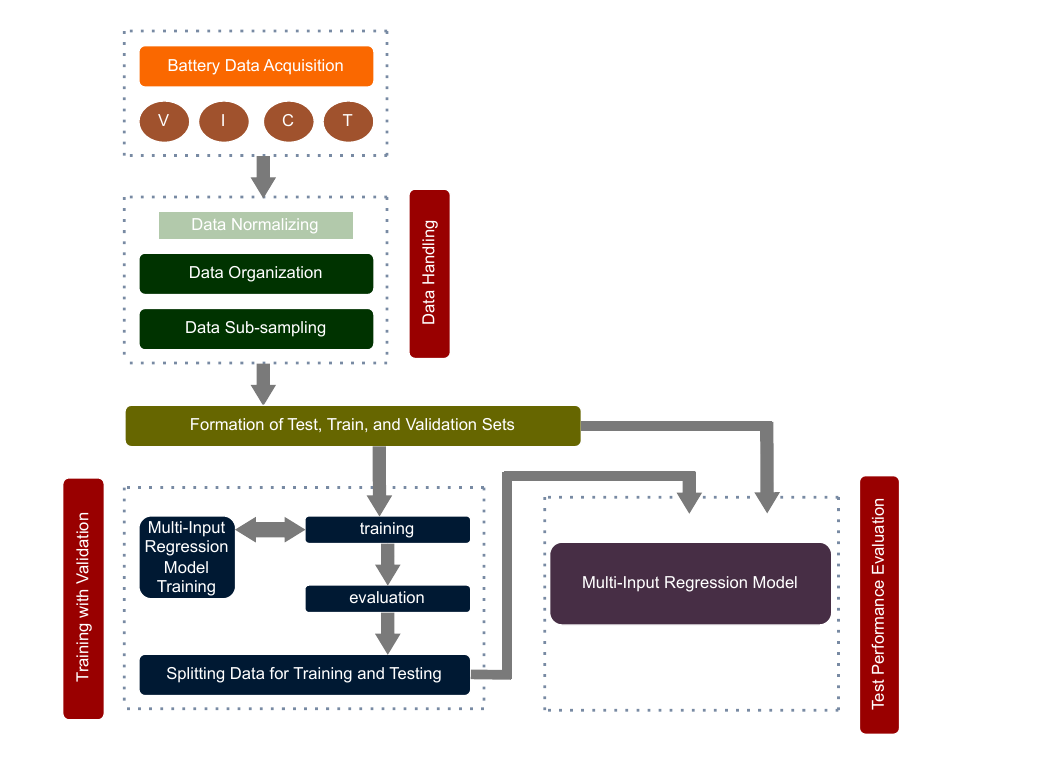}
    \caption{Pipeline for Multi-Input Regression Model in Battery State Estimation: Data Handling, Training, and Evaluation  }
    \label{test3}
\end{figure*}

Figure \ref{test3} illustrates a comprehensive pipeline for developing a multi-input regression model aimed at estimating battery states, specifically focusing on the SoC. The process is divided into three primary stages: \textit{Data Handling}, \textit{Training with Validation}, and \textit{Test Performance Evaluation}. As part of the Data Handling stage, raw battery data, including voltage (V), current (I), capacity (C), and temperature (T), is acquired and then processed through normalization, organization, and subsampling. Consequently, training, testing, and validation datasets are formed, ensuring data consistency, improving model robustness, and reducing computational complexity. As the multi-input regression model is developed through forward and backward propagation, the loss function is minimized iteratively through the Training with Validation stage. Optimum performance of the model is achieved by optimizing hyperparameters, including learning rate and batch size. Each subset plays a role in either training or validating over multiple iterations using cross-validation techniques to promote The final stage which is the Test Performance Evaluation involves evaluating the efficacy of the trained model on the test dataset. The metrics such as MSE, MAE and R2 score were measured to evaluate the predictive accuracy of the model. To determine the applicability for the real-time situations, the inference time and computational complexity were considered. It is possible to develop a robust and efficient multi-input regression model with the structured and multi-stage approach to estimate the battery state accurately and scale when necessary.
The BERT architecture of regression tasks is extended with a custom Bert Regression Model class. 768-dimensional output from the [CLS] token is extracted and moved through a linear layer. This process maps the output to a single scalar value which can match with MSE loss function. 
The training and evaluation are managed by the train and evaluate functions. While doing that, AdamW optimizer with MSE as the loss function is used. The training data is processed by the model for each epoch and the weights are updated according to the calculated loss. The model is evaluated on the test set following the training and the performance metrics for each epoch are recorded. 
Phyton’s logging module is used for logging operations. With that, data loading, training and evaluation steps are updated throughout the process.

\begin{figure}
    \graphicspath{ {D:\Stack} }
    \center \includegraphics[width=0.5\textwidth]{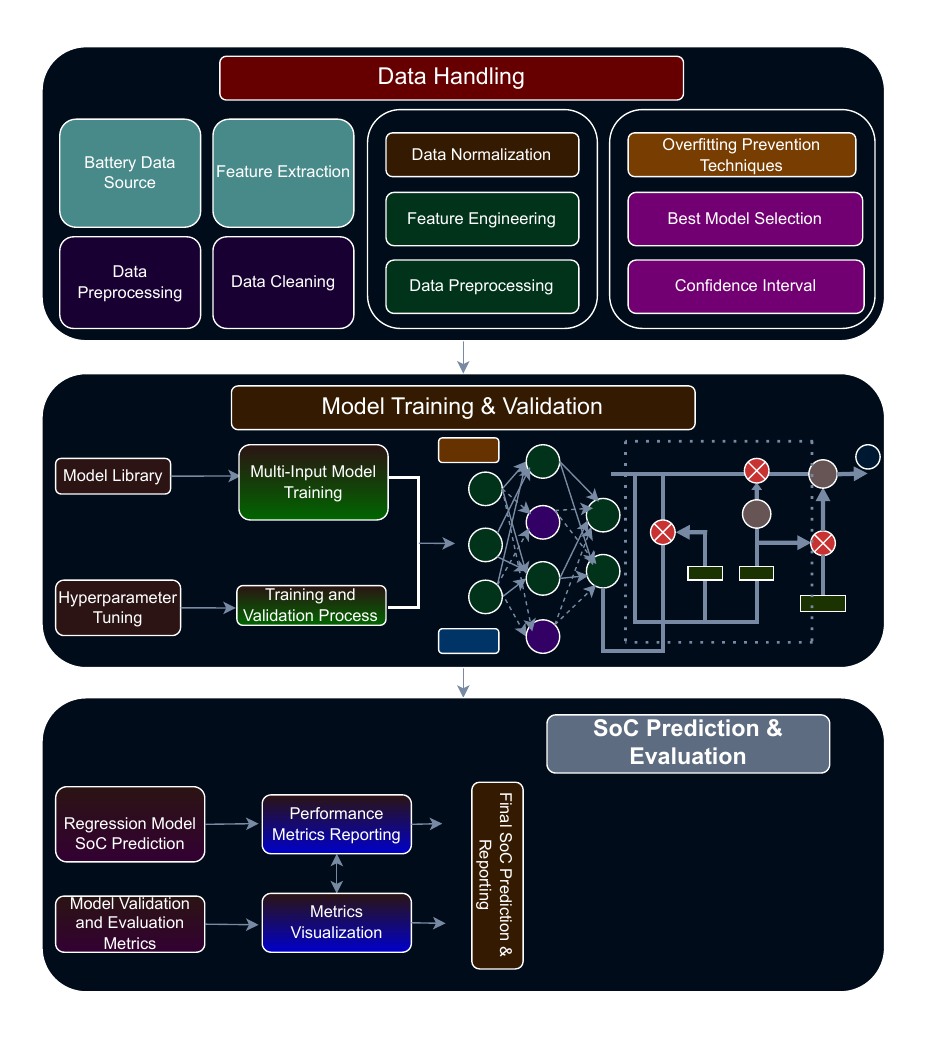}
    \caption{Pipeline for Multi-Input Regression Model Training and SoC Prediction with Comprehensive Data Handling and Evaluation  }
    \label{test4}
\end{figure}

Figure \ref{test4} shows the BERT's regression architecture is extended with the BertRegressionModel class. A linear layer is applied to the 768-dimensional output of the [CLS] token, converting it to a single scalar value compatible with the MSE loss function. Using the AdamW optimizer with MSE as the loss function, train\_and\_evaluate manage training and evaluation. Models process training data, compute predictions, and update weights according to calculated losses during each epoch. Model performance metrics are logged for each epoch after training on the test set. The logging module in Python provides updates on the loading, training, and evaluation of data. Hardware acceleration with CUDA is ensured by checking GPU availability.

Our framework leverages a transformer-based architecture with multi-head self-attention layers to capture intricate dependencies in battery degradation patterns. The model consists of:

\begin{itemize}
    \item \textbf{Input Embedding Layer}: Encodes cycle-based and instantaneous discharge data.
    \item \textbf{Multi-Head Self-Attention Mechanism}: Identifies long-term dependencies in battery data.
    \item \textbf{Feed-Forward Network}: Processes extracted features for robust regression.
    \item \textbf{Positional Encoding}: Retains sequential charge-discharge relationships.
    \item \textbf{Output Layer}: Generates SoC predictions using a fully connected regression head.
\end{itemize}

During the Data Handling phase, data is acquired from a variety of sources and preprocessed. This includes:

\begin{itemize}
    \item Data cleaning, normalization, and feature extraction to ensure data quality and consistency.
    \item Feature engineering techniques to derive informative variables and improve predictive performance.
    \item Techniques to enhance generalization across unseen data and prevent overfitting.
\end{itemize}

Multi-input regression models are optimized during the Model Training/Validation phase. To identify the best model architecture and parameters, hyperparameter tuning and various training methodologies are applied. This phase enables iterative updates while preventing overfitting by repeatedly splitting the training data into subsets. Model libraries allow systematic experimentation by storing different versions and configurations of models. A best model selection process ensures that only the most effective model is retained for deployment using statistical methods like confidence intervals.

While transformer models traditionally have high computational overhead, optimizations such as pruning, quantization, and TensorRT acceleration enable real-time inference on edge computing devices. Our approach justifies the use of transformers in EV applications through:

\begin{itemize}
    \item \textbf{Batch Inference}: Enables parallel processing of multiple SoC predictions efficiently.
    \item \textbf{Offline Pretraining, Online Inference}: Heavy computations occur during training, while inference is lightweight for real-time applications.
    \item \textbf{Hybrid Deployment Strategy}: Running SoC estimations at edge nodes (BMS units) while fine-tuning on cloud platforms.
\end{itemize}

The trained regression model is applied to test data to predict SoC. To evaluate predictive accuracy and robustness, MSE and MAE are calculated. Various performance metrics and visualization techniques provide insight into model performance. At the end of this phase, the best model is used to generate reliable SoC predictions, improving battery management and decision-making.

This structured approach to model development facilitates accurate and interpretable SoC prediction for battery management applications.

\section{Evaluation}

The performance of the BERT-based regression model was evaluated by using a held-out dataset. Additional metric and qualitative insights as well as the MSE were utilized for a thorough model performance apprehension. The model was primarily evaluated by the MSE. MSE calculates the average squared difference between predicted values and the actual values. The MSE is given by:

\begin{equation}
MSE = \frac{1}{n} \sum_{i=1}^{n} (y_i - \hat{y}_i)^2
\end{equation}

where \( y_i \) is the true value, and \( \hat{y}_i \) is the model’s predicted value. Lower MSE values reflect higher predictive accuracy. From this data, it is possible to understand that as MSE values become lower, the predictive accuracy increases. In addition to that, the R2 score is an indicator of the variance proportion obtained by the model in relation to a plain mean-based baseline.

The results obtained at the end of the fifth epoch included the following:
\begin{itemize}
    \item \textbf{Number of Batches Evaluated:} 122
    \item \textbf{Processing Speed:} 13.30 batches per second
    \item \textbf{Training Loss (MSE):} 655,290.2594
    \item \textbf{Test Loss (MSE):} 654,172.7254
    \item \textbf{Total Trainable Parameters:} 109,483,009
\end{itemize}

R\textsuperscript{2} score was negative, suggesting that the model underperformed compared to a simple mean baseline, despite the model's MSE decreasing over training epochs. As the test loss was high and the number of parameters was large, there was a significant risk of overfitting and difficulties generalizing to new data.

A substantial test loss of 654,172.7254 reveals significant deviations between model predictions and actual values. As a result of the high MSE, there were high prediction errors across samples, suggesting that the model was unable to capture accurate relationships.

This model evaluation was affected by a number of factors, which resulted in a high test loss. Initially designed for NLP tasks, the BERT-based architecture might not have been fully optimized for battery data regression without fine-tuning or domain-specific feature engineering. It is unlikely that the model would have been able to capture the intricate relationships required for accurate battery health predictions as a result. Furthermore, it appears that there were mismatches in features; the textual encoding approach may have failed to capture battery-related characteristics required for precise regression. It may have been difficult for the model to extract the appropriate representations of battery degradation patterns as a result. It is also possible that the model did not fully converge within the relatively short five-epoch training period. It may be beneficial to increase the number of epochs, along with adjusting learning rates or batch sizes, in order to support a more robust learning process.

Overfitting is further indicated by a gap between training and test loss. Hence, the model failed to generalize effectively to new, unseen samples, which reduced its predictive power. A number of improvements are recommended to meet these challenges. The model could learn more thoroughly if the training epochs were extended, capturing more complex relationships within the battery data. It is also possible to enhance the stability of the training process by optimizing hyperparameters, such as the learning rate and batch size. The use of regularization techniques, like dropouts or decay in weights, could mitigate the overfitting observed and enhance the model's ability to generalize. Additionally, a more refined approach to feature engineering, potentially integrating domain-specific features, could improve the quality of input data representation, enabling the model to capture essential trends in battery health. Exploring alternative model architectures that are better suited to regression tasks, such as hybrid converter-based approaches or architectures specifically optimized for regression, could yield better results in this area. These improvements would support more effective model training, reduce test loss, and potentially provide more accurate and generalizable predictions, enhancing the model’s practical applicability in battery health monitoring tasks.

\begin{table}[ht]
\caption{Performance Metrics of Multimodal LLM Framework on LTO Battery Capacity Fading}
\centering
\resizebox{\columnwidth}{!}{
\begin{tabular}{c|cc}
Methods/Metrics & MAE(\%) & Time \\
\hline
GPR & 21.00 & 34.50 \\
RD & 8.74 & 27.50 \\
SVR & 4.27 & 22.00 \\
CNN & 10.31 & 30.00 \\
\textbf{Our LLM} & \textbf{0.81} & \textbf{61.17} \\
\hline
\end{tabular}
}
\label{tab:battery_data_metrics_best}
\end{table}

Table \ref{tab:battery_data_metrics_best} compares performance metrics (MAE and processing time) of different methods on our battery dataset. Traditional methods like GPR and RD demonstrate moderate accuracy (MAE of 21.00\% for GPR and 8.74\% for RD) with relatively fast processing times (27.50–34.50 seconds). SVR achieves improved accuracy (MAE of 4.27\%) with even faster processing time (22.00 seconds), showing an efficient balance. CNN provides reasonable accuracy (MAE of 10.31\%) but is slightly slower (30.00 seconds). Our LLM framework achieves the highest accuracy (MAE of 0.81\%) but with a longer processing time (61.17 seconds), highlighting its superior precision for applications where accuracy is prioritized over speed.

\section{Conclusion}

This study demonstrates the potential of a transformer-based LLM framework for battery health monitoring and State of Health (SoH) estimation, expanding the application of BERT models into regression tasks beyond NLP. The model shows high accuracy in battery data prediction, as indicated by the low MAE values achieved in comparison with traditional methods like GPR, RD, SVR, and CNN. Despite its superior accuracy, the LLM framework incurs longer processing times, suggesting that it is best suited for applications where precision is more critical than real-time speed. For real-world integration, EV manufacturers and fleet operators can leverage this model for predictive maintenance, enabling early fault detection and optimizing battery replacement cycles. The ability to forecast SoH accurately allows for proactive interventions, reducing downtime and improving overall fleet efficiency. Regarding deployment strategies, the computational feasibility of LLM inference can be addressed through a hybrid approach. Edge computing devices can run lightweight inference models for real-time monitoring, while cloud-based processing can handle detailed periodic analyses, leveraging more computational resources. This dual-tier deployment ensures efficient use of computational power while maintaining the benefits of LLM-driven predictive analytics. However, the performance evaluation highlights challenges associated with using BERT-based models for battery data, particularly in processing efficiency. Future work should focus on optimizing model architecture, tuning hyperparameters, and implementing feature engineering strategies to balance accuracy with processing time, enhancing the model’s generalization and suitability for real-time deployment in battery management systems.

\section{Future Work}

Future efforts will focus on optimizing the LLM framework for faster processing, exploring feature engineering techniques, and refining model architecture to improve both accuracy and efficiency. Additionally, alternative transformer-based models and hybrid approaches will be investigated to enhance real-time prediction capabilities in battery management applications.

%\section{Acknowledgment}
%We acknowledge the Temsa Research R\&D Center for their generous financial support and the reviewers for their invaluable insights and suggestions that significantly contributed to the enhancement of our paper.

\bibliographystyle{IEEEtran}
\bibliography{external}

\end{document}